\documentclass[11pt,letterpaper]{article}
\usepackage[letterpaper]{geometry}
\usepackage{acl2012}
\usepackage{times}
\usepackage{latexsym}
\usepackage{amsmath}
\usepackage{amssymb}
\usepackage{multirow}
\usepackage{graphicx}
\usepackage{algorithm}
\usepackage{booktabs}
\usepackage{color}
\usepackage{subcaption}
\usepackage{afterpage}
\usepackage{anyfontsize}
\usepackage[noend]{algpseudocode}
\usepackage{url}
\makeatletter
\newcommand{\@BIBLABEL}{\@emptybiblabel}
\newcommand{\@emptybiblabel}[1]{}
\makeatother
\usepackage[hidelinks]{hyperref}

\newcommand{\imp}{\textcolor{black}}
\newcommand{\card}[1]{\lvert #1 \rvert}

\title{Unsupervised Learning of Morphological Forests}

\author{Jiaming Luo \\
 	CSAIL, MIT \\
  {\tt j\char`_luo@mit.edu} \\\And
  Karthik Narasimhan \\
  CSAIL, MIT \\
  {\tt karthikn@mit.edu} \\\And
  Regina Barzilay \\
  CSAIL, MIT \\
  {\tt regina@csail.mit.edu} }

\date{}

\begin{document}
\maketitle

\begin{abstract}
This paper focuses on unsupervised modeling of morphological families, collectively comprising a forest over the language vocabulary. This formulation enables us to capture edge-wise properties reflecting single-step morphological derivations, along with global distributional properties of the entire forest. These global properties constrain the size of the affix set and encourage formation of tight morphological families. The resulting  objective is solved using Integer Linear Programming (ILP) paired with contrastive estimation. We train the model by alternating between optimizing the local log-linear model and the global ILP objective. We evaluate our system on three tasks: root detection, clustering of morphological families and segmentation. Our experiments demonstrate that our model yields consistent gains in all three tasks compared with the best published results.\footnote{Code is available at \url{https://github.com/j-luo93/MorphForest}.}
\end{abstract}


\section{Introduction}

\begin{figure}
\centering
\includegraphics[width=1.05\linewidth]{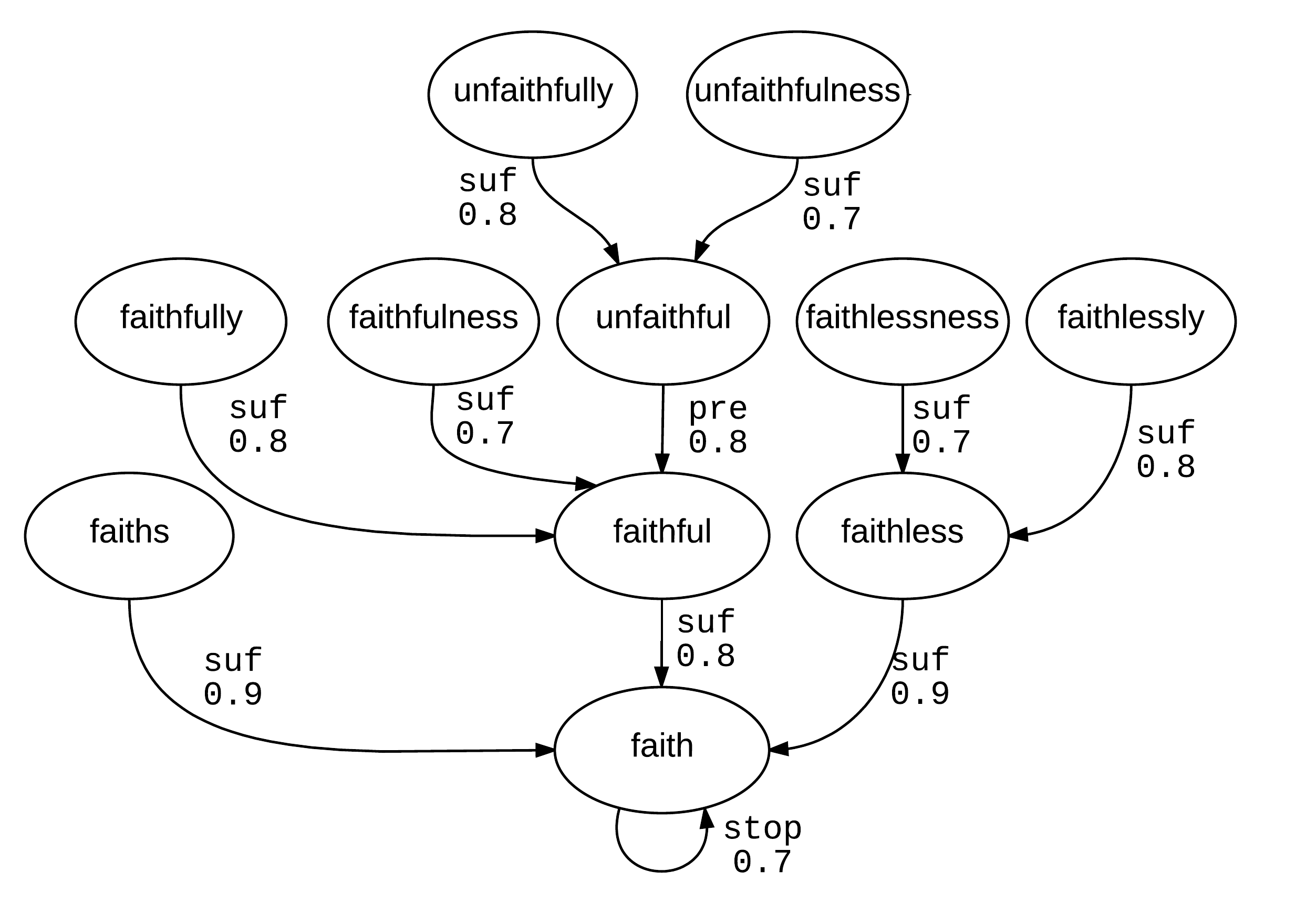}
\caption{An illustration of a single tree in a morphological forest. \textit{pre} and \textit{suf} represent prefixation and suffixation. Each edge has an associated probability for the morphological change.}
\label{fig:tree}
\end{figure}

The morphological study of a language inherently draws upon the existence of families of related words. All words within a family can be derived from a common root via a series of transformations, whether inflectional or derivational. Figure~\ref{fig:tree} depicts one such family, originating from the word \textit{faith}. This representation can benefit a range of applications, including segmentation, root detection and clustering of morphological families.

Using graph terminology, a full morphological assignment of the words in a language can be represented as a forest.\footnote{The correct mathematical term for the structure in Figure~\ref{fig:tree} is a \emph{directed 1-forest} or \emph{functional graph}. For simplicity, we shall use the terms forest and tree to refer to a directed 1-forest or a directed 1-tree because of the cycle at the root.} Valid forests of morphological families exhibit a number of well-known regularities. At the global level, the number of roots is limited, and only constitute a small fraction of the vocabulary. A similar constraint applies to the number of possible affixes, shared across families. At the local edge level, 
we prefer derivations that follow regular orthographic patterns and preserve semantic relatedness. We hypothesize that enforcing these constraints as part of the forest induction process will allow us to accurately learn morphological structures in an unsupervised fashion.

To test this hypothesis, we define an objective over the entire forest representation. 
The proposed objective is designed to maximize the likelihood of local derivations, while constraining the overall number of affixes and encouraging tighter morphological families.  We optimize this objective using integer linear programming (ILP), which is commonly employed to handle global constraints.  While in prior work, ILP has often been employed in supervised settings, we explore its effectiveness in unsupervised learning. We induce a forest by alternating between learning local edge probabilities using a log-linear model, and enforcing global constraints with the ILP-based decoder. With each iteration, the model progresses towards more consistent forests. 


We evaluate our model on three tasks: root detection, clustering of morphologically related families and segmentation. The last task has been extensively studied in the recent literature, providing us with the opportunity to compare the model with multiple unsupervised techniques. On benchmark datasets representing four languages, our model outperforms the baselines, yielding new state-of-the-art results.  For instance, we improve segmentation performance on Turkish by 4.4\% and on English by 3.7\%, relative to the best published results~\cite{narasimhan2015unsupervised}. Similarly, our model exhibits superior performance on the other two tasks. We also provide analysis of the model behavior which reveals that most of the gain comes from enforcing global constraints on the number of unique affixes.

\section{Related Work}
\label{sec:related_work}

\paragraph{Unsupervised morphological segmentation}
Most top performing algorithms for unsupervised segmentation today center around
modeling single-step
derivations~\cite{poon2009unsupervised,naradowsky2011unsupervised,narasimhan2015unsupervised}. A commonly used log-linear formulation enables these models to consider a rich set of features ranging from orthographic patterns to semantic relatedness. However, these models generally bypass global constraints \imp{\cite{narasimhan2015unsupervised} or require performing inference over very large spaces~\cite{poon2009unsupervised}}. As we show in our analysis (Section~\ref{sec:results}), this omission negatively affects model performance. 

In contrast, earlier work focuses on modeling global morphological assignment, using generative probabilistic
models~\cite{creutz2007unsupervised,snyder2008unsupervised,goldwater2009bayesian,sirts2013minimally}. These models are inherently limited in their ability to incorporate diverse features that are effectively utilized by local discriminative models. 

Our proposed approach attempts to combine the advantages of both approaches, by defining an objective that incorporates both levels of linguistic properties over the entire forest representation, and adopting an alternating training regime for optimization. 

\paragraph{Graph-based representations in computational morphology}
Variants of a graph-based representation have been used to model various morphological phenomena~\cite{dreyer2009graphical,peng-cotterell-eisner:2015:EMNLP,soricut2015unsupervised,TACL730}. The graph induction methods vary widely depending on the task and the available supervision. The distinctive feature of our work is the use of global constraints to guide the learning of local, edge-level derivations. 

\paragraph{ILP for capturing global properties}
Integer Linear Programming has been successfully employed to capture global constraints across multiple applications such as information extraction~\cite{roth2001relational}, sentence compression~\cite{clarke2008global}, and textual entailment~\cite{berant2011global}.  In all of these applications, the ILP formulation is used with a supervised classifier. Our work demonstrates that this framework continues to be effective in an unsupervised setting, providing strong guidance for a local, unsupervised classifier.

\section{Model}
\label{sec:model}



\label{sec:forest_rep}

Our model considers a full morphological assignment for all the words in a language, representing it as a forest.  Let $F = (V, E)$ be a directed graph where each word corresponds to a node $v\in V$. A directed edge $e = (v_c, v_p) \in E$ encodes a \emph{single} morphological derivation from a \emph{parent} word $v_p$ to a \emph{child} word $v_c$. Edges also reflect the type of the underlying derivation (e.g., prefixation), and an associated probability $\Pr(e)$. Note that the root of a tree is always marked with a self-directed (i.e. $v_c$ = $v_p$) edge associated
with the label \emph{stop}. Figure~\ref{fig:tree} illustrates a single tree in the forest.


\subsection{Inducing morphological forests}
\label{sec:objective}

We postulate that a valid assignment yields forests with the following properties: 

\begin{enumerate}
\item \textbf{Increased edge weights}\quad Edge weights reflect probabilities of single-step derivations based on the local features including orthographic patterns and semantic relatedness. This 
local information helps identify that the edge $(\textit{painter},\textit{paint})$ should be preferred over ($\textit{painter}, \textit{pain}$), because $-er$ is a valid suffix and \textit{paint} is semantically closer to \textit{painter}. 

\item \textbf{Minimized number of affixes}\quad Prior research has shown that 
local models tend to greatly overestimate the number of suffixes. For instance, the model of \newcite{narasimhan2015unsupervised} produces $617$ unique affixes when segmenting $10000$ English words. Thus, we explicitly encourage the model towards assignments with the least number of affixes.

\item \textbf{Minimized number of roots relatively to vocabulary size}\quad Similarly, the number of roots, and consequently the number of morphological families is markedly smaller than the size of the vocabulary. 
\end{enumerate}

The first property is local in nature, while the last two are global and embody the principle of Minimum Description Length (MDL).
Based on these properties, we formulate an objective function $\mathcal{S}(F)$ over a forest $F$:
\begin{equation}
\label{eq:raw_f}
\mathcal{S}(F) = -\frac{\sum_{e\in E}\log \Pr(e)}{\card{E}} + \alpha \card{\textit{Affix}} + \beta \frac{\card{F}}{\card{V}},
\end{equation}
where $\card{\cdot}$ denotes set cardinality, $\textit{Affix} = \{a_k \}_{k=1}^K$ is the set of all affixes, and $\card{F}$ is the number of trees in $F$.  $\card{E}$ and $\card{V}$ are the size of the edge set and vocabulary, respectively. The hyperparameters  $\alpha$ and $\beta$ capture the relative importance of the three terms. 

By minimizing this objective, we encourage assignments with high edge probabilities (first term), while limiting the number of affixes and morphological families (second and third terms, respectively). This objective can also be viewed as a simple log-likelihood objective regularized by the last two terms in Equation~(\ref{eq:raw_f}). 

\begin{figure}[t!]
	\captionsetup[sub]{labelformat=empty}
\centering
\frame{
\begin{subfigure}[(a)]{\linewidth}
	\centering
\includegraphics[width=\linewidth]{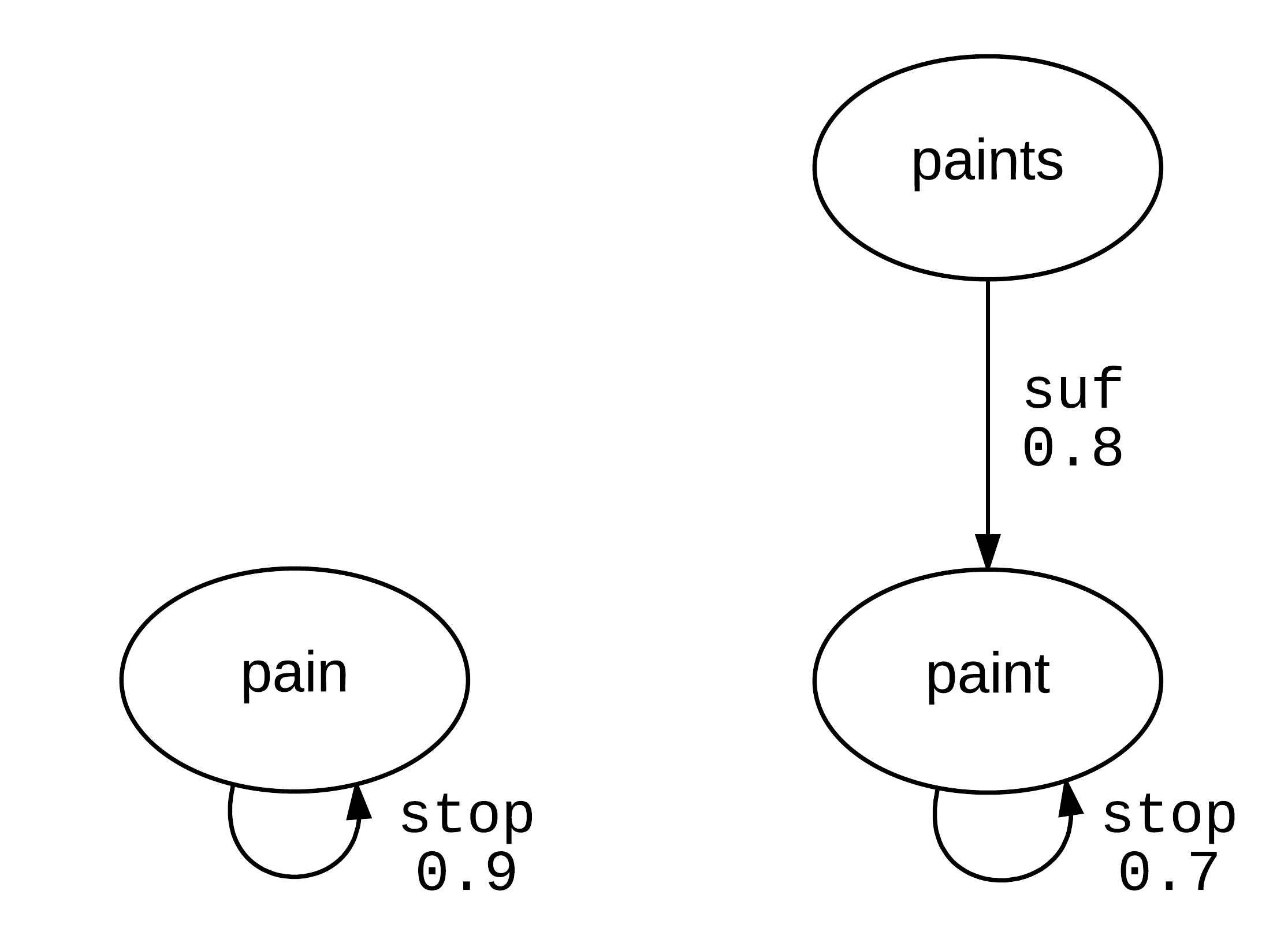}

\label{fig:candidate_a}
\end{subfigure}
}
\frame{
\begin{subfigure}[(b)]{\linewidth}
	\centering
\includegraphics[width=\linewidth]{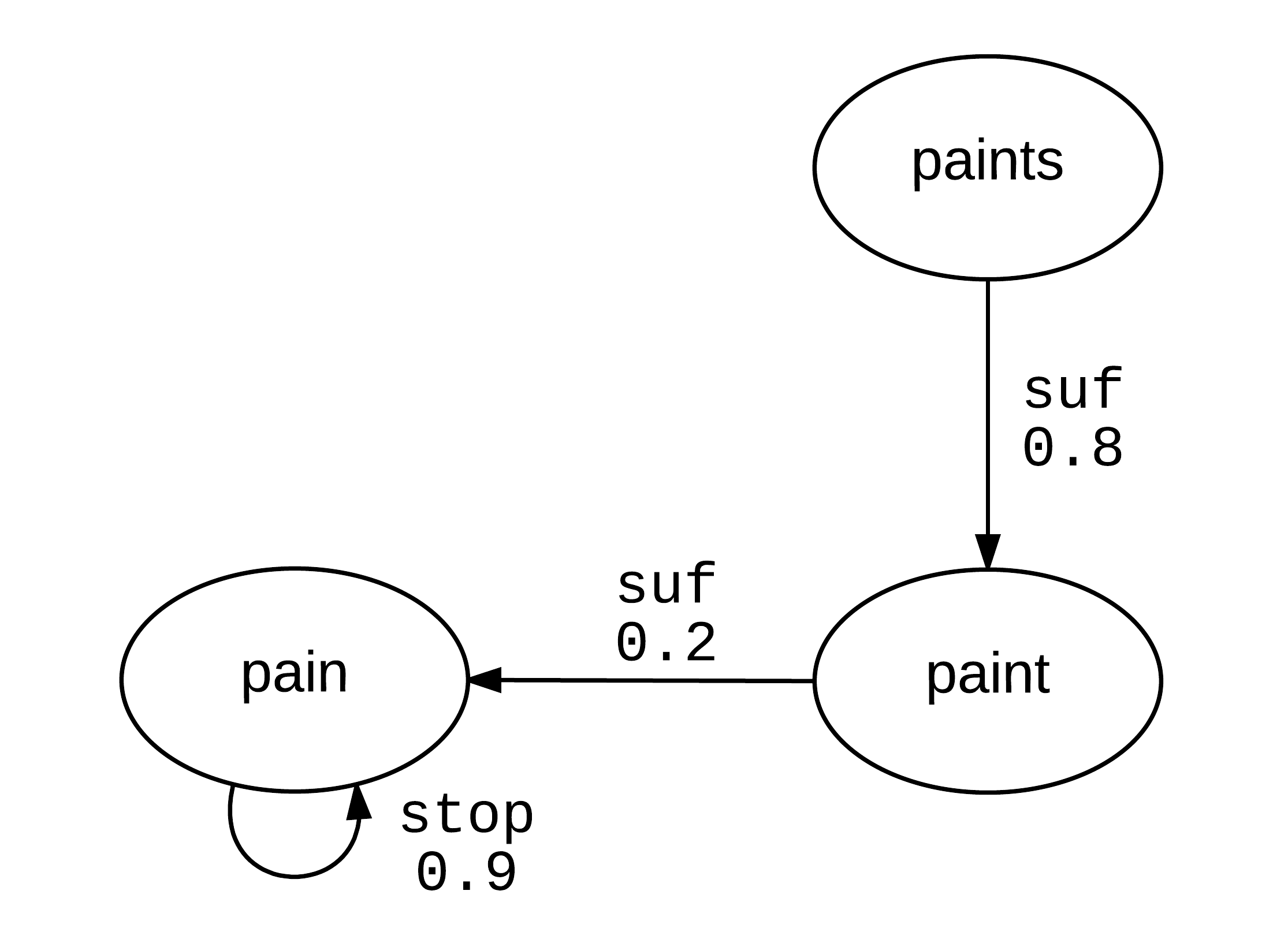}

\label{fig:candidate_b}
\end{subfigure}
}

\caption{Illustration of two chosen forest representations. The top forest has only one affix \textit{-s}, but two roots \{\textit{pain}, \textit{paint}\}. Shown in the bottom forest, choosing the edge $(\textit{paint}, \textit{pain})$ instead of $(\textit{paint}, \textit{paint})$ will introduce another affix \textit{-t}, while reducing the set of roots to just \{\textit{pain}\}.}
\label{fig:candidate}
\end{figure}

To illustrate the interaction between local and global constraints in this objective, consider an example in Figure~\ref{fig:candidate}. If the model selects a different edge -- e.g. (\emph{paint}, \emph{pain}) instead, all the terms in Equation~(\ref{eq:raw_f}) will be affected.

\subsection{Computing local probabilities}
\label{sec:compute_local}

We now describe how to parameterize $\Pr(e)$, which captures the likelihood of a single-step morphological derivation between two words. \imp{Following prior work~\cite{narasimhan2015unsupervised},} we model this probability using a log-linear model:
\begin{equation}
	\Pr(w, z) \propto \exp(\theta\cdot \phi(w, z)),
\end{equation}
where $\theta$ is the set of parameters to be learned, and $\phi(w, z)$ is the feature vector extracted from $w$ and $z$. Each candidate $z$ is a tuple (\emph{string}, \emph{label}), where \emph{label} refers to the label of the potential edge.

As a result, the marginal probability is
\begin{align*}
\Pr(w) &= \sum_{z \in C(w)} \Pr(w, z) \\
	   &=  \frac{\sum_{z \in C(w)} \exp(\theta\cdot\phi(w,z))}{\sum_{w'\in\Sigma^{*}} \sum_{z' \in C(w')} \exp(\theta\cdot\phi(w', z'))},
\end{align*}
where $\Sigma^{*}$ is the set of all possible strings. Computing the sum in the denominator is infeasible. Instead, we make use of \emph{contrastive estimation} \cite{smith2005contrastive}, substituting $\Sigma^*$ with a (limited) set of neighbor strings $N(w)$ that are orthographically close to $w$. This technique distributes the probability mass among neighboring words and forces the model to identify meaningful discriminative features. We obtain $N(w)$ by transposing characters in $w$, following the method described in \newcite{narasimhan2015unsupervised}. 

Now for the forest over the set of nodes $V$, the log-likelihood loss function is defined as:
\begin{align}
\mathcal{L}(V;\theta) &= - \sum_{v\in V} \log\Pr(v) \nonumber  \\
&= - \sum_{v\in V} \Big[ \log \sum_{z\in C(v)} \exp(\theta\cdot\phi(v, z)) \nonumber   \\
&\quad - \log \sum_{v'\in N(v)} \sum_{z' \in C(v')} \exp(\theta\cdot\phi(v', z')) \Big],
\label{eq:loss}
\end{align}

This objective can be minimized by gradient descent. 

\paragraph{Space of Possible Candidates}
We only consider assignments where the parent word is strictly shorter than the child to prevent cycles of length two or more. In addition to suffixation and prefixation, we also consider three types of transformations introduced in \newcite{goldwater2004priors}: repetition, deletion, and modification. We also handle compounding, where two stems are combined to form a new word (e.g., \textit{football}). One of these stems carries the main semantic meaning of the compound and is considered to be the parent of the word. \imp{Note that stems are not considered affixes, so this does not affect the affix list.}

We allow parents to be words outside $V$, since many legitimate word forms might never appear in the corpus. For instance, if we have $V =  \{ painter, paints \} $, the optimal solution would add an unseen word $paint$ to the forest, and choose edges $(\textit{painter}, \textit{paint})$ and $(\textit{paints}, \textit{paint})$.

\paragraph{Features}
We use the same set of features shown to be effective in prior work~\cite{narasimhan2015unsupervised}, including word vector similarity, beginning and ending character bigrams, word frequencies and affixes. Affix features are automatically extracted from the corpus based on string difference and are thresholded based on frequency. We also include an additional sibling feature that counts how many words are siblings of word $w$ in its tree. 
Siblings are words that are derived from the same parent, e.g., \textit{faithful} and \textit{faithless}, both from the word \textit{faith}.

\subsection{ILP formulation}
\label{sec:ILP}
Minimizing the objective in Equation~(\ref{eq:raw_f}) is challenging because the second and third terms capture discrete global properties of the forest, which prevents us from performing gradient descent directly. Instead, we formulate this optimization problem as Integer Linear Programming (ILP), where these two terms can be cast as constraints.\footnote{If we had prior knowledge of words belonging to the same family, we can frame the problem as growing a Minimum Spanning Tree (MST), and use Chu-Liu-Edmonds algorithm \cite{chu1965shortest,edmonds1967optimum} to solve it. However, this information is not available to us.}

For each child word $v_i \in V$, we have a bounded set of its candidate outgoing edges $C(v_i) = \{ z_{i}^{j} \} $, where $z_i^j$ is the $j$-th candidate for $v_i$. $C(v_i)$ is the same set as defined in Section~\ref{sec:compute_local}. Each edge is associated with $p_{ij}$, which is computed as $\log\Pr(z_{i}^{j} | v_i)$. Let $x_{ij}$ be a binary variable that has value 1 if and only if $z_{i}^{j}$ is chosen to be in the forest.
Without loss of generality, we assume the first candidate edge is always the self-edge (or \emph{stop} case), i.e., $z_{i}^1 = (v_i, \textit{stop})$. 
We also use a set of binary variables $\{ y_k \}$ to indicate whether affix $a_k$ is used at least once in $F$ (i.e. required to explain a morphological change).

Now let us consider how to derive our ILP formulation using the notations above.  Note that $\card{F}$ is equal to the number of self-edges $\sum_i x_{i1}$, and also a valid forest will satisfy $\card{V} = \card{E}$. Combining these pieces, we can rewrite the objective in equation (\ref{eq:raw_f}) and arrive at the following ILP formulation:
{
\small
\begin{align}
	&\underset{x_{ij}, y_k}{\text{minimize}}& -\frac{1}{\card{V}} &\sum_{ij} x_{ij }p_{ij} + \alpha \sum_k y_k + \frac{\beta}{\card{V}} \sum_i x_{i1} \nonumber \\
    &\text{subject to}&  x_{ij}, y_k &\in \{0, 1\}, \nonumber \\
    & & \sum_{j} x_{ij} &= 1, \forall i,  \label{constr:choose_1} \\
    & &  x_{ij} &\leq y_k, \text{if $a_k$ is involved in $z_{i}^{j}$}. \label{constr:affix}
\end{align}}%

Constraint~\ref{constr:choose_1} states that exactly one of the candidate edges should be chosen for each word. The last constraint implies that we can only consider this candidate (and construct the corresponding edge) when the involved affix\footnote{For English and German, where non-concatenative transformations are possible such as deletion of ending $e$ ($\textit{taking} \to \textit{take}$), we also include them in \textit{Affix}.} is used at least once in the forest representation. 



\subsection{Alternating training}
\label{sec:training}
The objective function contains two sets of parameters: a continuous weight vector $\theta$ that parameterizes edge probabilities, and binary variables $\{x_{ij}\}$ and $\{y_{k}\}$ in ILP. Due to the discordance between continuous and discrete variables, we need to optimize the objective in an alternating manner. Algorithm~\ref{alg} details the training procedure. After automatically extracting affixes from the corpus, we alternate between learning the local edge probabilities (line 3) and solving ILP (line 4). 

The feedback from solving ILP with the global constraints can help us refine the learning of local probabilities by removing incorrect affixes (line 5). For instance, automatic extraction based on frequencies can include \emph{-ers} as an English suffix. This is likely to be eliminated by ILP, since all occurrences of \emph{-ers} can be explained away without adding a new affix by concatenating \emph{-er} and \emph{-s}, two very common suffixes. 
After refining the affix set, we remove all candidates that involve any affix discarded by ILP. This corresponds to reducing the size of $C(w)$ in equation~(\ref{eq:loss}). We then train the log-linear model again using the newly-pruned candidate set. By doing so, we force the model to learn from better contrastive signals, and focus on affixes of higher quality, resulting in a new set of probabilities $\{p_{ij}\}$.
This procedure is repeated until no more affixes are rejected.\footnote{Typically the model converges after 5 rounds}

\begin{algorithm*}[th!]
  \caption{Morphological Forest Induction}\label{alg}
{\normalsize  
	\textbf{Input:}  wordlist $V$ \\
	\textbf{Output:} Forest representation of $V$ 
	\begin{algorithmic}[1]
    
    \State $\textit{Affix} \gets \textit{ExtractAffixes}(W)$
    \Comment{Extract common patterns as affixes from the wordlist}
		\For{$t\gets 1$ to $T$}
        		\Comment{Alternating training for $T$ iterations}
        		\State $p_{ij}^t \gets \textit{ContrastiveEstimation}(W, \textit{Affix})$ \Comment{Compute local probabilities, cf. Section~\ref{sec:compute_local} }
		 		\State $y^{*t}, F^t \gets \textit{ILP}(p^t_{ij})$ \Comment{Get indicators for affixes, and the forest, cf. Section~\ref{sec:ILP}}
                \State  \textit{PruneAffixSet}$(\textit{Affix}, y^{*t})$ 
\Comment{Prune affix set using the output from ILP, cf. Section~\ref{sec:training}}
        \EndFor
		\Return $F^T$			
	\end{algorithmic}
        }
\end{algorithm*}

\section{Experiments}
\label{sec:experiments}

We evaluate our model on three tasks: segmentation, morphological family clustering, and root detection. While the first task has been extensively studied in the prior literature,  we consider two additional tasks to assess the flexibility of the derived representation.

\subsection{Morphological segmentation}

\paragraph{Data}
We choose four languages with distinct morphological properties: English, Turkish, Arabic, and German. Our training data consists of standard datasets used in prior work. Statistics for all datasets are summarized in Table~\ref{tab:dataset}. Note that for the Arabic test set, we filtered out duplicate words, and we reran the baselines to obtain comparable results.

\imp{Following \newcite{narasimhan2015unsupervised}, we reduce the noise by truncating the training word list to the top $K$ frequent words.} In addition, we train word vectors~\cite{mikolov2013efficient} to obtain cosine similarity features. Statistics for all datasets are summarized in Table~\ref{tab:dataset}.

\begin{table}[t!]\centering
	\begin{tabular}{cccc}
		\multirow{2}{*}{\textbf{Language}} & \textbf{Train} & \textbf{Test} & \textbf{WordVec} \\
        & \textbf{\#Words} & \textbf{\#Words} & \textbf{\#Words} \\ 
        \toprule
        \multirow{2}{*}{English} & MC-10 & MC-05:10 & Wikipedia \\
        & 878K & 2212 & 129M\\
        \midrule 
        \multirow{2}{*}{Turkish} & MC-10 & MC-05:10 & BOUN \\
        & 617K & 2531 & 361M \\
        \midrule
        \multirow{2}{*}{Arabic} & Gigaword & ATB & Gigaword \\
        & 3.83M & 21085 & 1.22G \\
        \midrule
        \multirow{2}{*}{German} & MC-10 & Dsolve & Wikipedia \\
        & 2.34M & 15522 & 589M \\
	\end{tabular}
\caption{Data statistics: MC-10 = MorphoChallenge 2010 , MC:05-10 = aggregated from MorphoChallenge 2005-2010, BOUN = BOUN corpus~\protect\cite{sak2008turkish}, Gigaword = Arabic Gigaword corpus~\protect\cite{parker2011arabic}, ATB = Arabic Treebank~\protect\cite{maamouri2003arabic}. Duplicates in Arabic test set are filtered. Dsolve is the dataset released by~\protect\newcite{wurzner2015dsolve}, and for training German vectors, we use the pre-processed Wikipedia dump from~\protect\cite{polyglot:2013:ACL-CoNLL}.}
\label{tab:dataset}
\end{table}




\paragraph{Baselines}

We compare our approach against the state-of-the-art unsupervised method of \newcite{narasimhan2015unsupervised} which outperforms a number of alternative approaches~\cite{creutz2005inducing,virpioja2013morfessor,sirts2013minimally,lee2011modeling,stallard2012unsupervised,poon2009unsupervised}. For this baseline, we report the results of the publicly available implementation of the technique (\textit{NBJ'15}), as well as our own improved reimplementation (\textit{NBJ-Imp}). Specifically in \emph{NBJ-Imp}, we expanded the original algorithm to handle compounding, along with sibling features as described in Section~\ref{sec:compute_local}, making it essentially an ablation of our model without ILP and alternating training. We employ grid search to find the optimal hyperparameter setting.\footnote{$K \in \{ 2500, 5000, 10000 \}$, number of automatically extracted affixes $\in \{100, 200, 300, 400 , 500\}$}

We also include a supervised counterpart, which uses the same set of features as \textit{NBJ-Imp} but has access to gold segmentation during training (we perform 5-fold cross-validation using the same data). We obtain the gold standard parent-child pairs required for training from the segmented words in a straightforward fashion.

\paragraph{Evaluation metric}
Following prior work~\cite{virpioja2011tal}, we evaluate all models using the standard \textit{boundary precision and recall (BPR)}. This measure assesses the accuracy of individual segmentation points, producing IR-style \textit{Precision, Recall} and \textit{F1} scores. 

\paragraph{Training}
For unsupervised training, we use the gradient descent method \textsc{Adam}~\cite{kingma2014adam} and optimize over the whole batch of training words. We use a Gurobi\footnote{\url{http://www.gurobi.com/}} solver for the ILP.

\subsection{Morphological family clustering}
Morphological family clustering is the task of clustering morphologically related word forms. For instance, we want to group \textit{paint}, \textit{paints} and \textit{pain} into two clusters:
$\{\textit{paint, paints}\}$ and $\{\textit{pain}\}$. To derive 
clusters from the forest representation, we assume that all the words in the same tree form a cluster.

\paragraph{Data} To obtain gold information about morphological clusters, we use CELEX \cite{baayen1993celex}. Data statistics are summarized in Table~\ref{tab:data_clustering}. We remove words without stems from CELEX.\footnote{An example is \textit{aerodrome}, where both \textit{aero-} and \textit{drome} are affixes.} 

\paragraph{Baseline} We compare our model against \textit{NBJ-Imp} described above.
We select the best variant of our model and the base model based on their respective performance on the segmentation task.

\begin{table}
\centering
\begin{tabular}{cccc}
\multirow{2}{*}{\textbf{Language}} & \multirow{2}{*}{\textbf{\#Words}} & \multirow{2}{*}{\textbf{\#Clusters}} & \multirow{2}{*}\textbf{\textbf{\#Words}}\\
& & & \textbf{per Cluster} \\
\toprule
English & 75,416 & 20,249 & 3.72 \\
German & 367,967 & 28,198 & 13.05
\end{tabular}
\caption{Data statistics for the family clustering task (CELEX). We only evaluate on English and German, since these are the languages MorphoChallenge has segmentations for.}
\label{tab:data_clustering}
\end{table}

\paragraph{Evaluation}
We use the metrics proposed by \newcite{schone2000knowledge}. Specifically,
let $X_w$ and $Y_w$ be the clusters for word $w$ in our predictions and gold standard respectively. We compute the number of correct  ($\mathcal{C}$), inserted ($\mathcal{I}$) and deleted ($\mathcal{D}$) words for the clusters as follows:
\begin{align*}
\mathcal{C} &= \sum_{w \in W} \frac{\card{X_w \cap Y_w}}{\card{Y_w}} \\
\mathcal{I} &= \sum_{w \in W} \frac{\card{X_w \setminus Y_w}}{\card{Y_w}} \\
\mathcal{D} &= \sum_{w\in W} \frac{\card{Y_w \setminus X_w}}{\card{Y_w}}
\end{align*}
Then we compute $precision = \frac{\mathcal{C}}{\mathcal{C + I}}$, $recall = \frac{\mathcal{C}}{\mathcal{C + D}}$, $F1=2\frac{precision \cdot recall}{precision + recall}$.

\subsection{Root detection} 

In addition, we evaluate how accurately our model can predict the root of any given word. 

\paragraph{Data}
We report the results on the Chipmunk dataset \cite{cotterell2015labeled} which has been used for evaluating supervised models for root detection. Since our model is unsupervised, we report the performance both on the test set only, and on the entire dataset, combining the train/test split.  Statistics for the dataset are shown in Table~\ref{tab:data_root}.

\begin{table}
\centering
\begin{tabular}{ccc}
\multirow{2}{*}{\textbf{Language}} & \multirow{2}{*}{\textbf{\#Words}} & \textbf{\#Words} \\
& & \textbf{(Test only)} \\
\toprule
English & 1675&  687 \\
Turkish & 1759 & 763 \\
German & 1747 &749
\end{tabular}
\caption{Data statistics for root detection task. Duplicate words are removed.}
\label{tab:data_root}
\end{table}

\section{Results}
\label{sec:results}

In the following subsections, we report model performance on each one of the three evaluation tasks.
 
 \subsection{Segmentation}
 \begin{table}[!t]\centering
\small
\begin{tabular}{clccc}
	\multirow{2}{*}{\textbf{Language}} & \multirow{2}{1.7cm}{\centering \textbf{Method}} & \multicolumn{3}{c}{\textbf{BPR}} \\
    \cmidrule(lr){3-5}
    & & \textbf{P} & \textbf{R} & \textbf{F} \\
    \toprule
    \multirow{7}{*}{English} & \emph{Supervised} & 0.905 & 0.813 & 0.856 \\
    & \textit{NBJ'15} & 0.807  & 0.722 & 0.762 \\
    & \textit{NBJ-Imp} & 0.820 & 0.726 & 0.770  \\
    \cmidrule(lr){2-5}
    & \emph{Our model} & 0.838 & 0.729 & 0.780  \\
    & \emph{ + Sibl}  & 0.796 & 0.739 & 0.767\\
    & \emph{ + Comp}  & 0.840 & 0.761 & \textbf{0.799}\rlap{$^*$}\\
    & \emph{ + Comp, Sibl}  & 0.815 & 0.774 & 0.794  \\
    \midrule\midrule
    \multirow{7}{*}{Turkish} & \emph{Supervised} &  0.826 & 0.803 & 0.815 \\
    & \textit{NBJ'15} & 0.743 & 0.520 & 0.612  \\
    & \emph{NBJ-Imp} & 0.697 & 0.583 & 0.635  \\
    \cmidrule(lr){2-5}
    & \emph{Our model} & 0.717 & 0.577 & 0.639  \\
    & \emph{ + Sibl}  & 0.698 & 0.619 & \textbf{0.656}\rlap{$^*$}  \\
    & \emph{ + Comp}  & 0.716 & 0.581 & 0.642 \\
    & \emph{ + Comp, Sibl}  & 0.692 & 0.621 & 0.655  \\
    \midrule\midrule
    \multirow{7}{*}{Arabic} & \emph{Supervised}  & 0.904 &  0.921 &  0.912 \\

    & \textit{NBJ'15} & 0.840 & 0.724 & 0.778 \\
    & \emph{NBJ-Imp} & 0.866 & 0.725 & 0.789   \\
    \cmidrule(lr){2-5}
    & \emph{Our model} & 0.848 & 0.769 & 0.806  \\
    & \emph{ + Sibl}  & 0.829 & 0.787 & \textbf{0.807}\rlap{$^*$}  \\ 
    & \emph{ + Comp}  & 0.851 & 0.765 & 0.806  \\
    & \emph{ + Comp, Sibl}  & 0.881 & 0.745 & \textbf{0.807}\rlap{$^*$} \\
    \midrule\midrule 
        \multirow{7}{*}{German\footnotemark} 
 & \emph{Supervised} & 0.823 & 0.810 &  0.816 \\

	& \textit{NBJ'15}   &  0.716 & 0.275 & 0.397 \\
    & \emph{NBJ-Imp} & 0.790 & 0.480 & 0.597  \\

    \cmidrule(lr){2-5}
     & \emph{Our model} & 0.774	& 0.540	& 0.636\\
    & \emph{ + Sibl}  & 0.711 &	0.514 &	0.596 \\
    & \emph{ + Comp}  & 0.777	& 0.595 & \textbf{0.674}\rlap{$^*$} \\
    & \emph{ + Comp, Sibl}  & 0.701 &	0.616 &	0.656 \\
    \bottomrule\bottomrule
\end{tabular}
\caption{Segmentation results for the supervised model and three unsupervised models: the state-of-the-art system \textit{NBJ'15}~\protect\cite{narasimhan2015unsupervised}, our improved implementation of their system \textit{NBJ-Imp} and our model.  For our model, we also report results with different feature combinations. \textit{+ Sibl} and \textit{+ Comp} refer to addition of sibling and compounding features respectively. 
Best hyperparameter values for unsupervised baselines (\emph{NBJ'15}, \emph{NBJ-Imp}) are chosen via grid search, 
while for our model, we use 10K words and top $500$ affixes throughout.
* implies statistical significance with $p < 0.05$ against the \textit{NBJ-Imp} model using the sign test~\protect\footnotemark.}
\label{tab:results}
\end{table}
\addtocounter{footnote}{-1} 
\footnotetext{We used cosine similarity features in all experiments. But the root forms of German verbs are rarely used, except in imperative sentences. Consequently they barely have trained word vectors, contributing to the low recall value. We suspect better treatment with word vectors can further improve the results.}
\stepcounter{footnote}
\footnotetext{\url{http://www.mathcracker.com/sign-test.php}}

From Table~\ref{tab:results}, we observe that our model consistently outperforms the baselines on all four languages. Compared to \textit{NBJ'15}, our model has a higher F1 score by $3.7\%, 4.4\%, 2.9\%$ and  $27.7\%$ on English, Turkish, Arabic and German, respectively. While the improved implementation~\textit{NBJ-Imp} benefits from the addition of compounding and sibling features, our model still delivers an absolute increase in F1 score, ranging from $1.8\%$  to $7.7\%$ over \textit{NBJ-Imp}. 
Note that our model achieves higher scores even without tuning the threshold $K$ or the number of affixes, whereas the baselines have optimal hyperparameter settings via  grid search. 

\begin{figure}[!t]
\centering
\includegraphics[width=\linewidth]{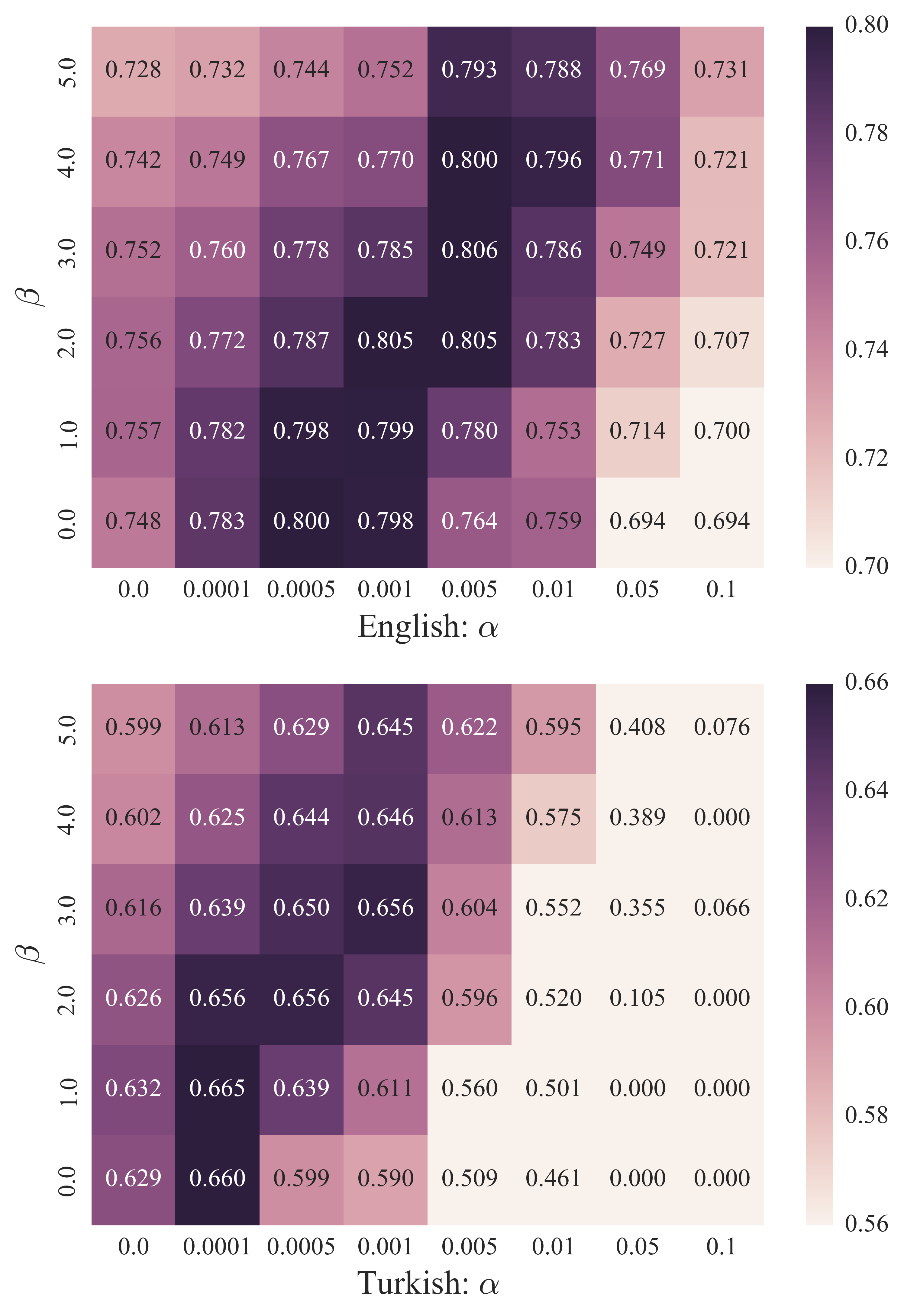}
\caption{Heat maps of $\alpha$ and $\beta$ for English and Turkish. Darker cells mean higher scores. Models used are $+ Comp$ for English and $+ Sibl$ for Turkish. }
\label{fig:hyper}
\end{figure}

To understand the importance of global constraints (the last two terms of equation~\ref{eq:raw_f}), we analyze our model's performance with different values of $\alpha$ and $\beta$ (see Figure~\ref{fig:hyper}). The first constraint, which controls the size of the affix set, plays a more dominant role than the second. By setting $\alpha=0.0$, the model scores at best $75.7\%$ on English and $63.2\%$  on Turkish, lower than the baseline. While the value of $\beta$ also affects the F1 score, its role is secondary in achieving optimal performance.   

The results also demonstrate that language properties can greatly affect the feature set choice. For fusional languages such as English, computing of sibling features is unreliable. For example,
two descendants of the same parent \textit{spot} -- \textit{spotless} and \textit{spotty} -- may not be necessarily identified as such by a simple sibling computation algorithm, since they undergo different changes.  In contrast, Turkish is highly agglutinative, with minimal (if any) transformations, but each word can have up to hundreds of related forms. Consequently, sibling features have different effects on English and Turkish, leading to changes of $-0.3\%$ and $+2.1\%$ in F1 score respectively. 

\begin{figure}[!t]
\centering
\includegraphics[width=1.0\linewidth]{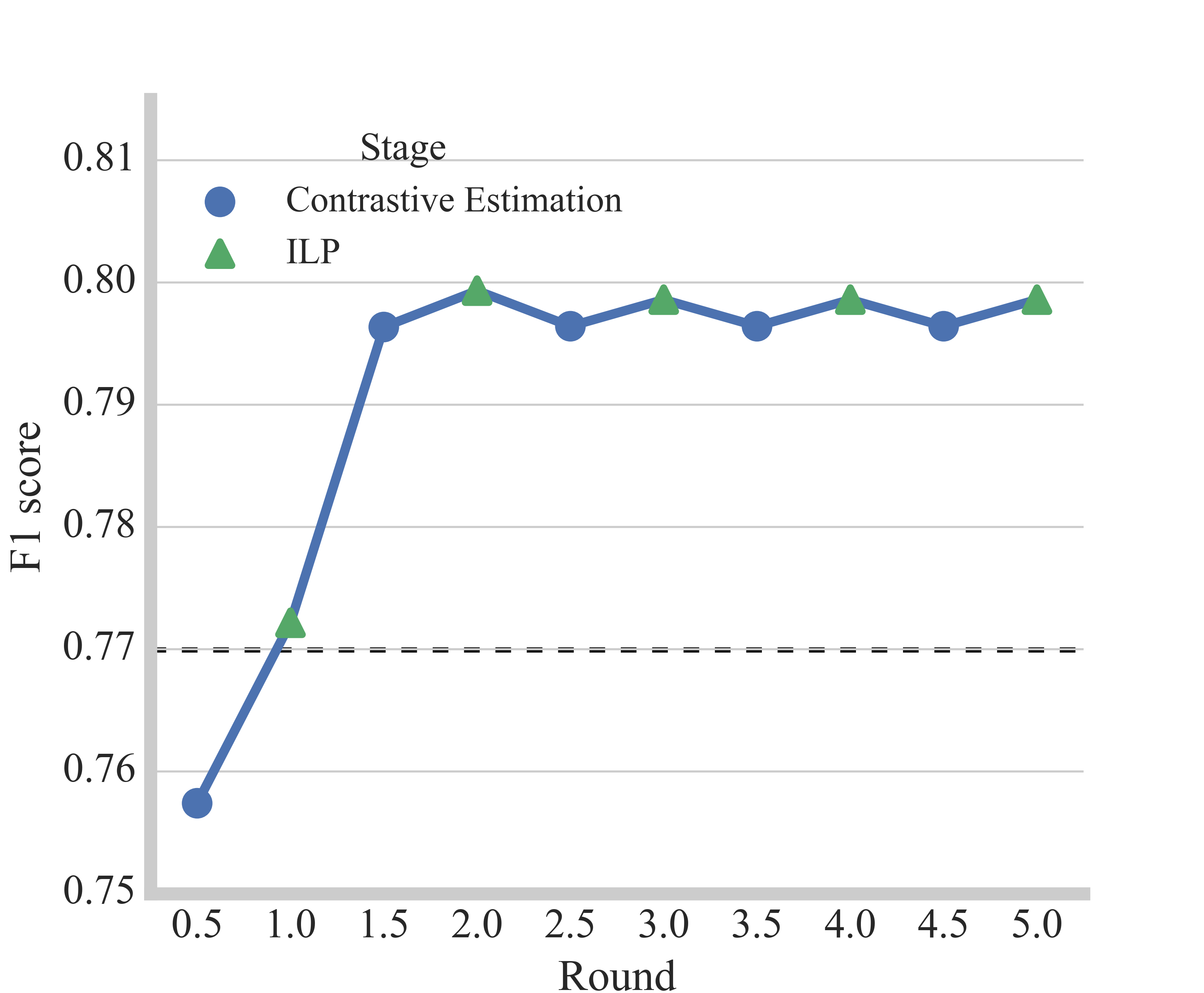}
\caption{F1 score vs round of training, for \textit{+ Comp} on English. Training log-linear models and solving ILP are marked by circles and triangles respectively. Best result for \textit{NBJ-Imp}  is represented as a dashed horizontal line. }
\label{fig:round}
\end{figure}

\paragraph{Understanding model behavior}
We find that much of the gain in model performance comes from the first two rounds of training. As Figure~\ref{fig:round} shows, the improvement mainly stems from solving ILP in the first round, followed by training the log-linear model in the second round after removing affixes and pruning candidate sets. This is exactly what we expect from the ILP formulation -- to globally adjust the forest by reducing the number of unique affixes. We find this to be quite effective -- in English, out of $500$ prefixes, only $6$ remain:  \textit{de}, \textit{dis}, \textit{im}, \textit{in}, \textit{re}, and \textit{un}. Similarly, only $72$ out of $500$ suffixes survive after this reduction. 

\paragraph{Robustness}
We also investigate how robust our model is to the choice of hyperparameters. Figure~\ref{fig:hyper} illustrates that we can obtain a sizable boost over the baseline by choosing $\alpha$ and $\beta$ within a fairly wide region. Note that $\alpha$ takes on a much smaller value than $\beta$, to maintain the two constraints ($|\textit{Affix}|$ and $\frac{|F|}{|V|}$) at comparable magnitudes.

\begin{figure}[!t]
\centering
\includegraphics[width=0.9\linewidth]{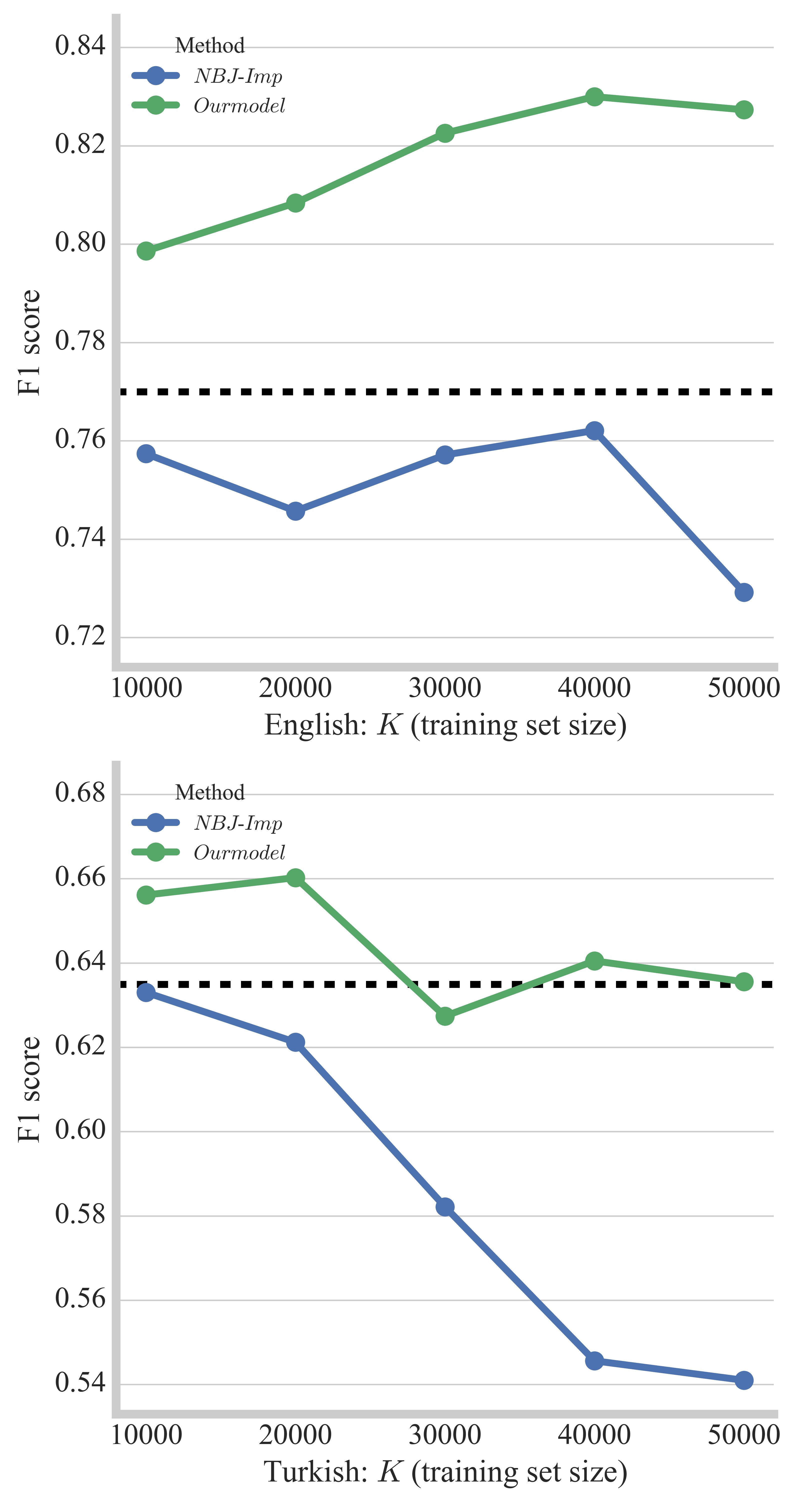}
\caption{Performance using bigger training sets. $+ Comp$ for English and $+ Sibl$ for Turkish. Dashed lines represent the best results for \textit{NBJ-Imp} (with smaller training sets).}
\label{fig:big}
\end{figure}

\newcite{narasimhan2015unsupervised} observe that after including more than $K = 10000$ words, the performance of the unsupervised model drops noticeably. \imp{In contrast, our model  handles training noise more robustly, resulting in a steady boost or not too big drop in performance with increasing training size} (Figure~\ref{fig:big}). In fact, it scores $83.0\%$ with $K=40000$ on English, a $\textbf{6.0\%}$ increase in absolute value over the baseline.

\paragraph{Qualitative analysis}
Table~\ref{tab:positive} shows examples of English words that our model segments correctly, while \emph{NBJ'15} fails on them. We present them in three categories (top to bottom) based on the component of our model that contributes to the successful segmentation. 
The first category benefits from a refinement of affix set, by removing noisy ones, such as \textit{-nce, -ch, } and \textit{k-}. This leads to correct stopping as in the case of \emph{knuckle} or induction of the right suffix, as in \emph{divergence}. Further, a smaller affix set also leads to more concentrated weights for the remaining affixes. For example, the feature weight for \textit{-ive} jumps from $0.06$ to $0.25$, so that the derivation $\textit{negative}\to\textit{negate}$ is favored, as shown in the second category. Finally, the last category lists some compound words that our model successfully segments.

\begin{table}[!t]
\centering
\begin{tabular}{ll}
\textbf{NBJ-Imp} & \textbf{Our model} \\
\toprule
\textit{diverge-nce} & \textit{diverg-ence} \\
 \textit{lur-ch} & \textit{lurch} \\
\textit{k-nuckle} & \textit{knuckle} \\
 \midrule
\textit{negative}  & \textit{negat-ive} \\
\textit{junks} & \textit{junk-s} \\
\textit{unreserved} & \textit{un-reserv-ed} \\
\midrule
\textit{gaslight-s} & \textit{gas-light-s} \\
 \textit{watercourse-s} & \textit{water-course-s} \\
 \textit{expressway} & \textit{express-way}\\
\end{tabular}
\caption{Some English words that our model segments correctly which the unsupervised base model (\emph{NBJ'15}) fails at.}
\label{tab:positive}
\end{table}

\subsection{Morphological family clustering}
We show the results for morphological family clustering in Table~\ref{tab:results_clustering}. For both languages, our model increases $precision$ by a wide margin, with a modest boost for $recall$ as well. This corroborates our findings in the segmentation task, where our model can effectively remove incorrect affixes while still encouraging words to form tight, cohesive families. 

\begin{table}
\begin{tabular}{clccc}
\textbf{Language} & \textbf{Method} & \textbf{P} & \textbf{R} & \textbf{F} \\
\toprule
\multirow{2}{*}{English} & \textit{NBJ-Imp} & 0.328 & 0.680 & 0.442 \\
& \textit{Our model} & 0.895 & 0.715 & \textbf{0.795} \\
\midrule
\multirow{2}{*}{German} & \textit{NBJ-Imp} & 0.207 & 0.421 & 0.278 \\
& \textit{Our model} & 0.471 & 0.484 & \textbf{0.477} \\
\end{tabular}
\centering
\caption{Results for morphological family clustering. P = precision, R = recall. }
\label{tab:results_clustering}
\end{table}

\subsection{Root detection}
Table~\ref{tab:results_root} summarizes the results for the root detection task. Our model shows consistent improvements over the baseline on all three languages. We also include the results on the test set of two \emph{supervised} systems: \textit{Morfette} \cite{Chrupala2008LearningMW} and \textit{Chipmunk} \cite{cotterell2015labeled}. \textit{Morfette} is a string transducer while \textit{Chipmunk} is a segmenter. Both systems have access to morphologically annotated corpora.

Our model is quite competitive against \textit{Morfette}. In fact, it achieves higher accuracy for English and Turkish. Compared with \textit{Chipmunk}, our model scores 0.65 versus 0.70 on English, bridging the gap significantly. However, the high accuracy for morphologically complex languages such as Turkish and German suggests that unsupervised root detection remains a hard task.

\begin{table}
\centering
\begin{tabular}{clcc}
\multirow{2}{*}{\textbf{Language}} & \multirow{2}{*}{\textbf{Method}} & \multirow{2}{*}{\textbf{Accuracy}}  & \textbf{Accuracy}\\
& &  & \textbf{(Test only)} \\
\toprule
\multirow{4}{*}{English} & \textit{NBJ-Imp} & 0.590 & 0.595 \\
& \textit{Our model} & \textbf{0.636}  & \textbf{0.649} \\ 
\cmidrule(lr){2-4}
& \textit{Morfette} & - & 0.628 \\
& \textit{Chipmunk} & - & 0.703  \\
\midrule
\multirow{4}{*}{Turkish} & \textit{NBJ-Imp}  & 0.446 & 0.442\\
&\textit{Our model}& \textbf{0.463} & \textbf{0.467} \\
\cmidrule(lr){2-4}
& \textit{Morfette}& - & 0.268 \\
& \textit{Chipmunk} & - & 0.756\\
\midrule
\multirow{4}{*}{German} & \textit{NBJ-Imp}& 0.347 & 0.331\\
& \textit{Our model}& \textbf{0.383} & \textbf{0.364}\\
\cmidrule(lr){2-4}
& \textit{Morfette} & - & 0.438 \\
& \textit{Chipmunk} & -  & 0.674\\
\end{tabular}
\caption{Results for root detection. Numbers for \textit{Morfette} and \textit{Chipmunk} are reported by \protect\newcite{cotterell2015labeled}.}
\label{tab:results_root}
\end{table}

\section{Conclusions}

In this work, we focus on unsupervised modeling of morphological families, collectively defining a forest over the language vocabulary. This formulation enables us to incorporate both local and global properties of morphological assignment. The resulting objective is solved using Integer Linear Programming (ILP) paired with contrastive estimation. Our experiments demonstrate that our model yields consistent gains in three morphological tasks compared with the best published results.


\section*{Acknowledgement}
We thank Tao Lei, Yuan Zhang and the members of the MIT NLP group
for helpful discussions and feedback. We are also grateful to
anonymous reviewers for their insightful comments.
\bibliography{references}
\bibliographystyle{acl2012}

\end{document}